\documentclass[conference]{IEEEtran}
\IEEEoverridecommandlockouts
\usepackage{cite}
\usepackage{amsmath,amssymb,amsfonts}
\usepackage{algorithmic}
\usepackage{graphicx}
\usepackage{textcomp}
\usepackage{xcolor}
\usepackage{booktabs}
\usepackage[caption=false,font=footnotesize]{subfig}
\usepackage{placeins}
\usepackage{float} 
\usepackage{url}
\usepackage[T1]{fontenc}
\usepackage{newtxtext,newtxmath}
\begin{document}

\title{A Comparative Evaluation of Teacher-Guided Reinforcement Learning Techniques for Autonomous Cyber Operations}

\author{\IEEEauthorblockN{Konur Tholl}
\IEEEauthorblockA{\textit{Royal Military College of Canada} \\
\textit{Electrical and Computer Engineering}\\
Kingston, Canada \\
konur.tholl@rmc-cmr.ca
}
\and
\IEEEauthorblockN{Mariam El-Mezouar}
\IEEEauthorblockA{\textit{Royal Military College of Canada} \\
\textit{Department of Mathematics}\\
Kingston, Canada \\
mariam.el-mezouar@rmc-cmr.ca}
\and
\IEEEauthorblockN{Ranwa Al Mallah}
\IEEEauthorblockA{\textit{Polytechnique Montreal} \\
\textit{Computer and Software Engineering}\\
Montreal, Canada \\
ranwa.al-mallah@polymtl.ca}
}

\maketitle

\begin{abstract}
Autonomous Cyber Operations (ACO) rely on Reinforcement Learning (RL) to train agents to make effective decisions in the cybersecurity domain. However, existing ACO applications require agents to learn from scratch, leading to slow convergence and poor early-stage performance. While teacher-guided techniques have demonstrated promise in other domains, they have not yet been applied to ACO. In this study, we implement four distinct teacher-guided techniques in the simulated CybORG environment and conduct a comparative evaluation. Our results demonstrate that teacher integration can significantly improve training efficiency in terms of early policy performance and convergence speed, highlighting its potential benefits for autonomous cybersecurity.
\end{abstract}


\section{Introduction}
The scale and capabilities of offensive cyber operations are substantial, and constantly evolving. It is not practical for humans to manually defend their systems against this ever-evolving attack space. As such, Autonomous Cyber Operations (ACO) was created to enable agents to make effective decisions on behalf of humans. Current ACO applications rely on Reinforcement Learning (RL) to train these agents, allowing them to learn by directly interacting with an environment. However, these agents start from scratch and learn solely from the environment's signals. This ultimately leads to poor initial performance, as the agent must execute undesirable actions and learn from the resulting reward signals. 

We propose addressing this limitation by integrating a teacher into the RL pipeline \cite{tholl_thesis_2025}. Specifically, we implement four distinct teacher-guided techniques that incorporate a pretrained RL agent, which the learning agent can leverage to improve decision-making. We validate the techniques using CybORG's Cage Challenge 2 environment \cite{baillie_cyborg_2020, cagechallenge2github}. For clarity, CybORG Cage Challenge 2 will be referred to as CybORG for the remainder of this paper.

Specifically, our contributions are as follows:

\begin{itemize}
    \item \textit{Teacher-Guided Implementations}. We implement four distinct teacher-guided RL techniques and adapt them to the CybORG environment. These include feature space modification, reward shaping, action masking, and auxiliary loss, each leveraging a pretrained RL agent as the teacher.

    \item \textit{Comprehensive Evaluation}. We evaluate the performance of the four teacher-guided methods, measuring their impact on early training performance, convergence speed, and final policy quality using CybORG.

    \item \textit{Practical Insights for ACO}. We demonstrate that incorporating a pretrained teacher into the RL pipeline can significantly improve early-stage performance and accelerate learning. Our results highlight the potential of teacher-guided RL as a step toward improving agents' training efficiency for ACO.
\end{itemize}


\section{Background}
Existing work in ACO leverages RL to train agents to autonomously make decisions on behalf of people in the cybersecurity domain \cite{wiebe_learning_2023, mcdonald_competitive_2023, loevenich_designllmcyborg_2024, baillie_cyborg_2020, Faizan_MARL_2024}. Unlike traditional Machine Learning (ML) approaches, which require vast datasets, RL enables agents to learn through direct interaction with an environment \cite{sutton_reinforcement_2014}.

Current ACO applications initialize these agents as untrained models that learn solely from the environment's signals \cite{wiebe_learning_2023,Faizan_MARL_2024,loevenich_designllmcyborg_2024, mcdonald_competitive_2023}. In these settings, there is no external source they can leverage to help make decisions. Agents must carry out unfavorable actions and learn from the consequences, leading to training inefficiencies. Additionally, this approach is especially problematic in ACO, given the critical nature of the cybersecurity domain, where poor actions can have dire consequences.

\section{Teacher-Guided Techniques}
The idea of incorporating a teacher into the RL pipeline to augment decision-making is not novel in itself, and several methods for teacher integration have been proposed in prior work. In our work, we use Proximal Policy Optimization (PPO) as the RL algorithm for all implementations \cite{schulman_proximal_2017}. 

Teacher integration initially used generic imitation approaches, such as the one proposed by M. Pfeiffer et al., where the teacher generates a dataset that is used to train an RL agent prior to interacting with the environment  \cite{pfeiffer_reinforcedimitationlearning_2018}. However, in addition to needing to create a dataset that captures all the edge cases, this approach uses the teacher's feedback in isolation of the environment. If there is any misalignment, the agent will have to ``unlearn'' the teacher's policy during the transition to RL, leading to longer training times. 

\subsection{Reward Shaping}
To address these potential limitations, A. Beikmohammadi and S. Magnusson incorporated the teacher's feedback through reward shaping, where the teacher computed its own reward signal that was used alongside the environment's reward to train the agent \cite{beikmohammadi_ta-explorerewardshaping_2023}. To facilitate a smooth transition from teacher-guided to independent RL, they gradually decreased the teacher's contribution to the reward signal as training progressed:

\begin{equation}
    R^e(s_t, a_t, s_{t+1}) = 
    \beta(e) R^A + 
    (1 - \beta(e)) R^T
\end{equation}

where \(R^A\) is the environment's original reward signal, \(R^T\) is the reward computed from the teacher and \(\beta(e)\) is the impact of the teacher's reward signal, which is gradually increased, thereby giving less weight to the teacher.

\begin{figure}[!t]
    \centering
    \includegraphics[width=\columnwidth]{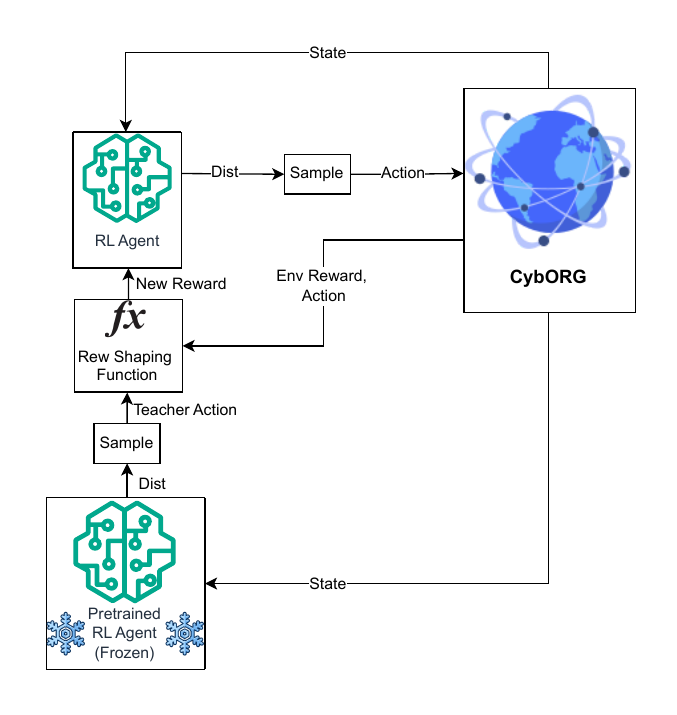}
    \caption{Overview of the reward shaping process. The teacher modifies the reward signal used to train the agent.}
    \label{fig:rewardshaping}
\end{figure}

For our implementation within CybORG, we increased the reward signal if the agent selected the teacher's recommended action or a host pertaining to the recommended action. In particular:
\begin{equation}
    \label{eq:rewardshape}
    r_{t}(a)=
\begin{cases} 
      r_{envt}(a) + c_{1}, & \text{if } a = a^{T} \text{ (recom action)} \\
      r_{envt}(a) + c_{2}, & \text{if } a \in A^{T} \text{ (recom host)}\\
      r_{envt}(a), & \text{otherwise}
\end{cases}
\end{equation}

where \(r_{envt}(a)\) is the original reward, and \(c_{1}\) and \(c_{2}\) are both positive floats with \(c_{1}>c_{2}\). We kept the teacher's influence constant before abruptly ceasing it for one implementation and gradually decayed it for the other. Fig. \ref{fig:rewardshaping} provides a high-level overview of how we incorporated the teacher's guidance using reward shaping.

\subsection{Action Masking}
Z. Wang et al. proposed a more direct approach than reward shaping, where the agent's probability distribution is directly modified based on the teacher's recommendation \cite{wang_learningactionmasking_2024}. They employed a fixed binary mask in which the probabilities of non-recommended actions were set to 0.

For our implementation, we decreased the probability of selecting any action not recommended by the teacher. In particular:

\begin{equation}
    \label{eq:mask1}
    \pi_{masked \theta}(a_{t})=\pi_{\theta}(a_{t})*M_{t}(a_{t})
\end{equation}
where \(\pi_{masked \theta}(a_{t})\) and \(\pi_{\theta}(a_{t})\) are the masked and original policies, respectively, and:
\begin{equation}
\label{eq:mask2}
M_{t}(a)=
\begin{cases} 
      1, & \text{if } a \in A_{\text{T}} \text{ (recommendations)} \\
      c_{3}, & \text{otherwise}
\end{cases}
\end{equation}

Similar to reward shaping, we gradually decayed the teacher's influence by increasing \(c_{3}\) every training interval until it reached 1 and kept \(c_{3}\) constant before removing the teacher's guidance (setting \(c_{3}\) to 1). 

\begin{figure}[!b]
    \centering
    \includegraphics[width=\columnwidth]{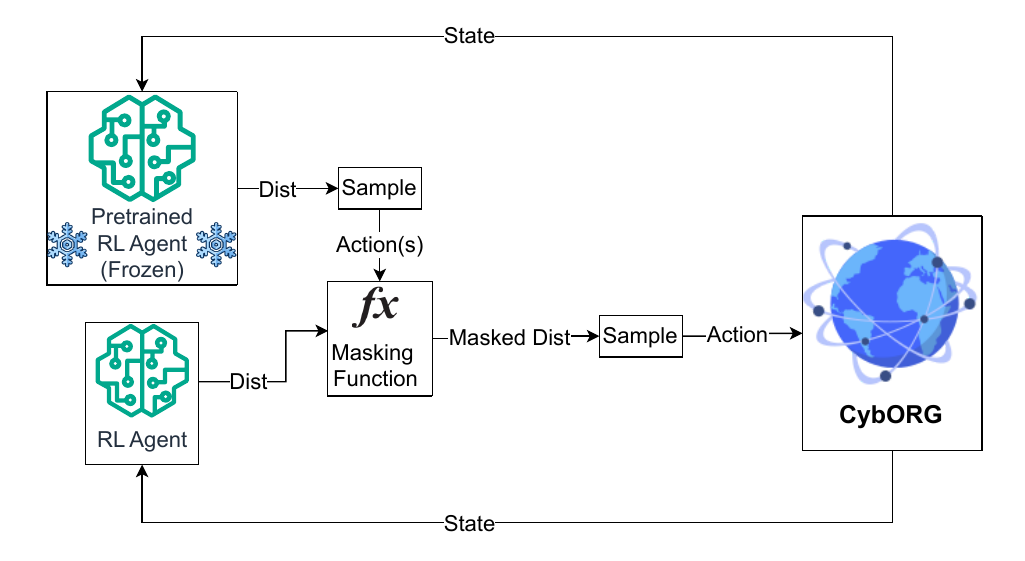}
    \caption{Illustration of how action masking was employed. The teacher's recommendations are used to modify the agent's probability distribution prior to sampling an action.}
    \label{fig:actionmasking}
\end{figure}

We present an overview of our action masking process in Fig. \ref{fig:actionmasking}.

\subsection{Auxiliary Loss}
Z. Zhou et al. demonstrated another way to incorporate a teacher by directly modifying the agent's loss function \cite{zhou_llm4rl_2024}:

\begin{equation}
    L^{tot}(\theta)=L^{A}(\theta)+\mathbf{\lambda}L^{T}(\theta)
\end{equation}
where \(L^{A}\) is the PPO agent's actor loss, \(L^{T}\) is the computed teacher loss, and \(\mathbf{\lambda}\) is the impact of the teacher, which is gradually decreased.

Similarly, we focused on incorporating the auxiliary loss signal for the PPO agent's actor network. To facilitate a smoother transition from teacher-guided to independent RL, we scaled the teacher's loss by a scalar and added an entropy term:

\begin{equation}
\label{eq:auxloss1}
\begin{aligned}
L^{tot}(\theta) &= \sigma * L^{A}(\theta) 
+ (1-\sigma) * L^{Teacher}(\theta) \\
&\quad + c_{4} S\!\left(\pi_{\theta}(\cdot \mid s_{t})\right)
\end{aligned}
\end{equation}

where \(\sigma\) is inversely proportional to the teacher's impact on the overall loss and proportional to the original PPO loss. \(S(\pi_{\theta}(\cdot|s_{t}))\) is the actor's entropy, representing the randomness within its policy, and \(c_{4}\) controls the extent to which this contributes to the loss, with a higher value encouraging exploration. Finally, \(L^{Teacher}(\theta)\) is the teacher's loss computed as the log probability of selecting its recommended action in the agent's current policy:

\begin{equation}
\label{eq:auxloss2}
L^{Teacher}(\theta) = -log\pi_{\theta}(a_{t}^{Teacher}|s_{t})
\end{equation}

\begin{figure}[!b]
    \centering
    \includegraphics[width=\columnwidth]{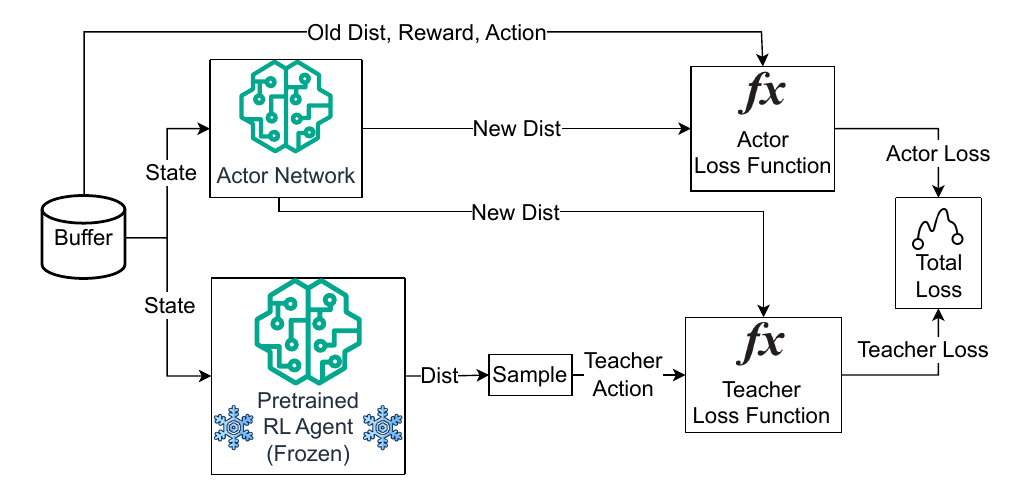}
    \caption{Overview of the auxiliary loss process. The teacher's recommendation was used to directly compute the loss applied to the actor network.}
    \label{fig:auxiliaryloss}
\end{figure}

We performed two implementations of auxiliary loss: gradually increasing \(\sigma\) to stabilize the transition from teacher-guided learning, and abruptly removing it to validate the teacher's robustness in PPO. For entropy, we gradually increased \(c_{4}\) during the teacher-guided phase to maximize exploration immediately following the teacher's guidance, thereby increasing the chances of surpassing the teacher. Once transitioned, we gradually decayed \(c_{4}\) to encourage convergence onto an optimal policy. Fig. \ref{fig:auxiliaryloss} illustrates how we incorporated the auxiliary loss signal.

\subsection{Feature Space Modification}
The discussed techniques incorporate the teacher's feedback after the agent has produced a policy, whether indirectly through reward shaping or directly through action masking. 

J. Wang et al. proposed a novel method that incorporates the teacher's guidance by modifying the agent's input \cite{wang_boostinginstruccomprehension_2025}. In particular, they leverage the teacher to decompose the environment's state into simpler sub-tasks for the agent to follow. 

Rather than using the teacher to simplify the environment's state space, we appended the teacher's recommendation as an additional feature to enable the agent to make an informed decision. In particular, we update the agent's feature space as:

\begin{equation}
\label{eq:featspacemod}
s_{t}=[s_{ti},\text{encoded}(a_{t}^{\text{Teacher}})]
\end{equation}

where \(s_{ti}\) is the environment's state and \(\text{encoded}(a_{t}^{\text{Teacher}})\) is the encoded teacher's recommendation. We trialed three different encodings for the teacher's recommendation: a binary value, a one-hot encoding, and a normalized float. For each of these, the features pertaining to the teacher's recommendation remained between 0 and 1, ensuring the agent did not overemphasize actions represented by a higher number due to their greater influence on the gradient.

\begin{figure}[!b]
    \centering
    \includegraphics[width=\columnwidth]{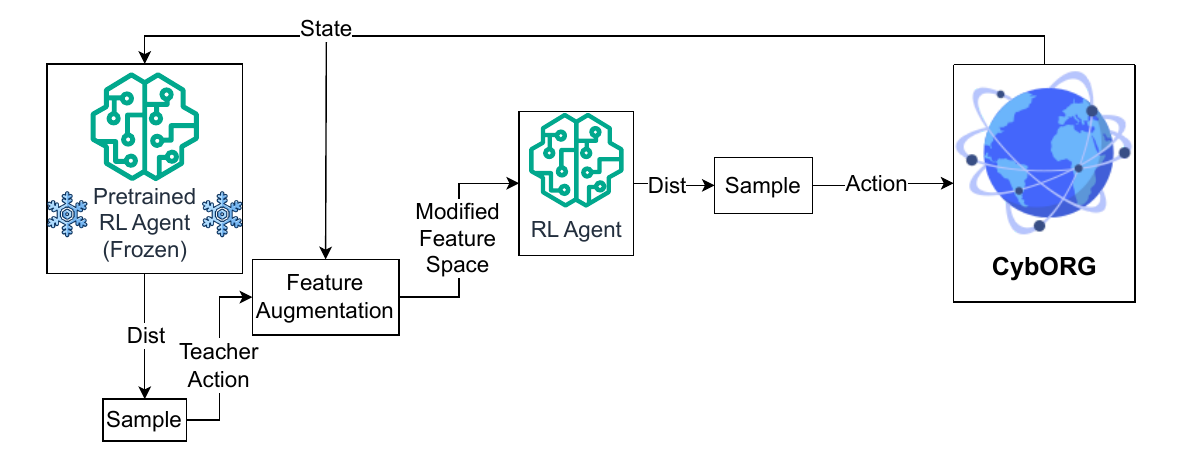}
    \caption{Overview of feature space modification. The teacher's guidance is appended to the agent's state space as an additional feature to help with decision-making.}
    \label{fig:featurespacemod}
\end{figure}

Fig. \ref{fig:featurespacemod} illustrates how we used the teacher to modify the agent's state space.

\section{Experimental Setup}
We evaluated each teacher-guided technique against the baseline PPO agent using identical hyperparameters; the only difference was the teacher integration. Each evaluation was carried out using the mean across 10 independent runs of 500 episodes for each technique. We used Standard Error (SE) instead of Standard Deviation (SD) to measure variance, as we are comparing two independent techniques instead of the discrepancy between individual runs; however, both are valid ways to measure variance.

We used a pretrained RL agent trained for 100 episodes as the teacher for each technique. This duration was chosen to make the teacher effective for early training while allowing both the guided-agent and baseline to surpass the teacher in a reasonable amount of time.

For evaluating the success of the individual techniques, we focused on three criteria: the performance of the final policy, the number of timesteps required to converge to a favorable policy, and the initial performance of the policy.

For feature space modification in particular, we used Local Interpretable Model Explanations (LIME) to quantify the weight of individual features on the agent's decision, allowing us to assess the agent's reliance on teacher-recommended features \cite{ribeiro_lime_2016}. To facilitate this, checkpoints of the actor network were saved at episodes 1, 8, 16, 50, 100, 200, 300 and 500. From here, perturbations of an identical state were used to estimate the impact each feature had on the agent's decision for each checkpoint.

\section{Evaluation}
In this section, we present, interpret, and compare the results of the discussed teacher-guided techniques. 

\subsection{Reward Shaping}
For the reward shaping implementation discussed in \eqref{eq:rewardshape}, we added a reward of \(c_1\)=2.5 to the environment's reward signal if the agent selected the action recommended by the teacher, and a reward of \(c_2\)=1.0 if the agent selected an action that pertained to the host recommended by the teacher. These rewards were abruptly halted at episode 40 (training interval 5). We also decayed them by a factor of 10\% every training interval to facilitate a smoother transition to independent RL.

\begin{figure}[!b]
    \centering
    \includegraphics[width=\columnwidth]{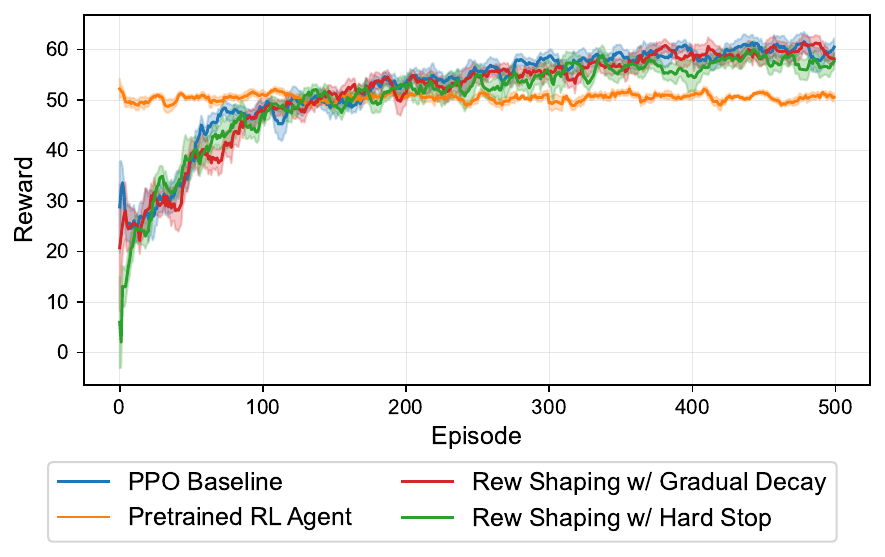}
    \caption{Comparing reward shaping against the PPO baseline across 10 independent runs of 500 episodes with a running average of 10. The shaded regions represent a ±1 SE. For fair comparison, this figure includes the unmodified rewards for the teacher-guided techniques.}
    \label{fig:rewardshapingeval}
\end{figure}

We present the results of the reward shaping technique in Fig. \ref{fig:rewardshapingeval}.

As shown, both implementations of our reward shaping technique exhibit no noticeable improvements from the PPO baseline, with all three plots converging to a reward of approximately 60 by episode 500.

\subsection{Action Masking}
For the masking implementations, we directly modified the agent's policy to decrease the likelihood of selecting any action not recommended by the teacher as shown in \eqref{eq:mask1} and \eqref{eq:mask2}. We used a masking value of \(c_3\)=0 during the first training interval and increased it by 25\% per interval thereafter. We also applied a hard stop masking approach, where we kept the masking value at \(c_3\)=0 and abruptly ceased it after four training intervals.

Furthermore, we trialed an additional technique, where the masking was applied to a subset of actions that pertain to the teacher's recommended host. For this implementation, we started with a mask of \(c_3\)=0, and applied the same transitional techniques where we gradually decayed the teacher's guidance and abruptly stopped it. For the gradual decay, we decreased the teacher's influence by 10\% every training interval and for the hard stop, we completely removed the masking after six training intervals.

\begin{figure}[!b]
    \centering
    \includegraphics[width=\columnwidth]{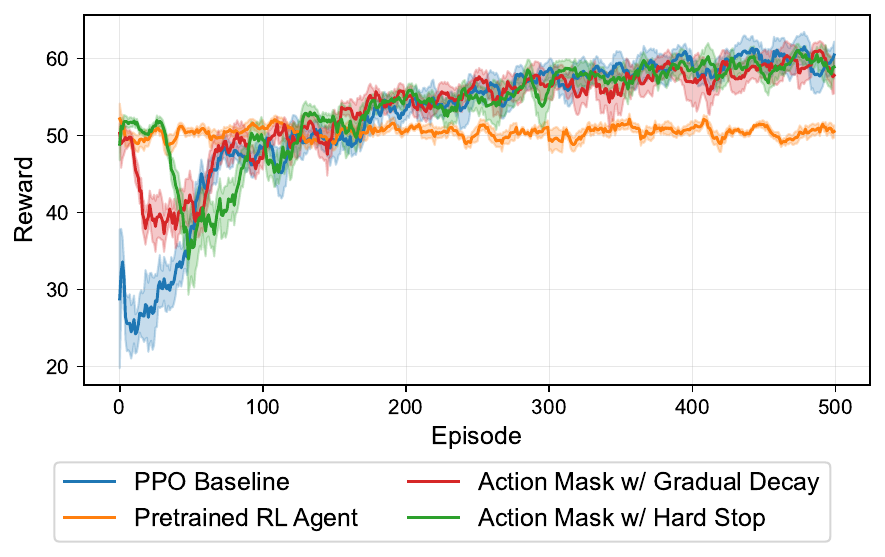}
    \caption{Comparing action masking against the PPO baseline across 10 independent runs of 500 episodes with a running average of 10. The shaded regions represent a ±1 SE.}
    \label{fig:actionmaskingeval}
\end{figure}

\begin{figure}[!t]
    \centering
    \includegraphics[width=\columnwidth]{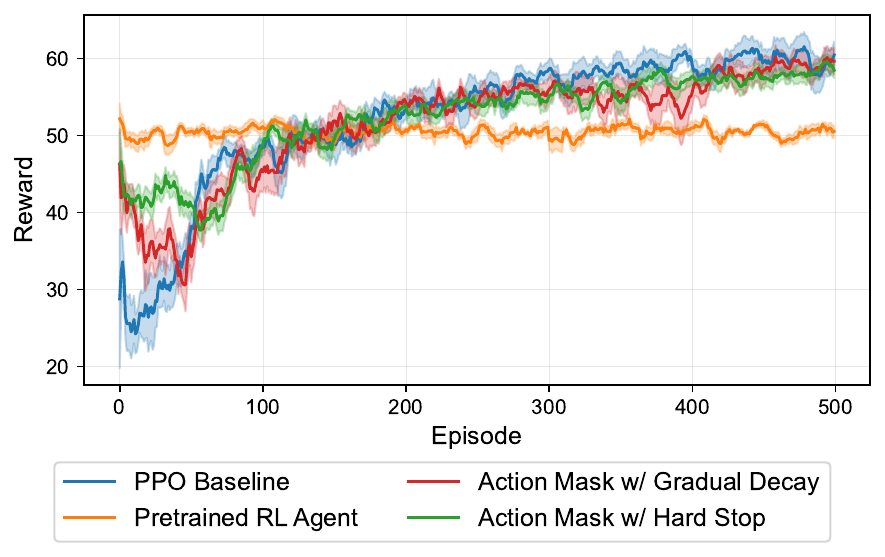}
    \caption{Comparing host masking against the PPO baseline across 10 independent runs of 500 episodes with a running average of 10. The shaded regions represent a ±1 SE.}
    \label{fig:hostmaskingeval}
\end{figure}

Figs. \ref{fig:actionmaskingeval} and \ref{fig:hostmaskingeval} compare the action and host masking techniques against the baseline. The results show that the initial performance is superior during the masking phase, but there is a notable drop during the transition to independent RL. Allowing the RL agent to explore within a subset of actions yields lower initial performance than single action masking, but has a relatively lower drop in performance during the transition to independent RL. Overall, the gradually decayed action masking provides the best balance of initial performance and transition to independent RL, with initial rewards of approximately 50 and the policy dropping to a lowest value of approximately 40 as the teacher's influence is decayed.

\subsection{Auxiliary Loss}
We incorporated the teacher's guidance as an auxiliary loss signal for the PPO agent's actor network, as described in \eqref{eq:auxloss1} and \eqref{eq:auxloss2}. The actor's loss was initially computed solely from the teacher's recommendation and gradually decayed by 25\% each training interval thereafter. For the hard stop implementation, we removed the teacher's influence completely after three training intervals.

To encourage the agent to surpass the teacher, we increased the entropy coefficient (\(c_{4}\)) by \(5e^{-4}\) every teacher-guided training interval so that it prioritized exploration as it approached independent RL. Once the teacher's guidance was removed, we decreased \(c_{4}\) by \(2e^{-4}\) every interval until it reached its starting value of 0.005.

\begin{figure}[!b]
    \centering
    \includegraphics[width=\columnwidth]{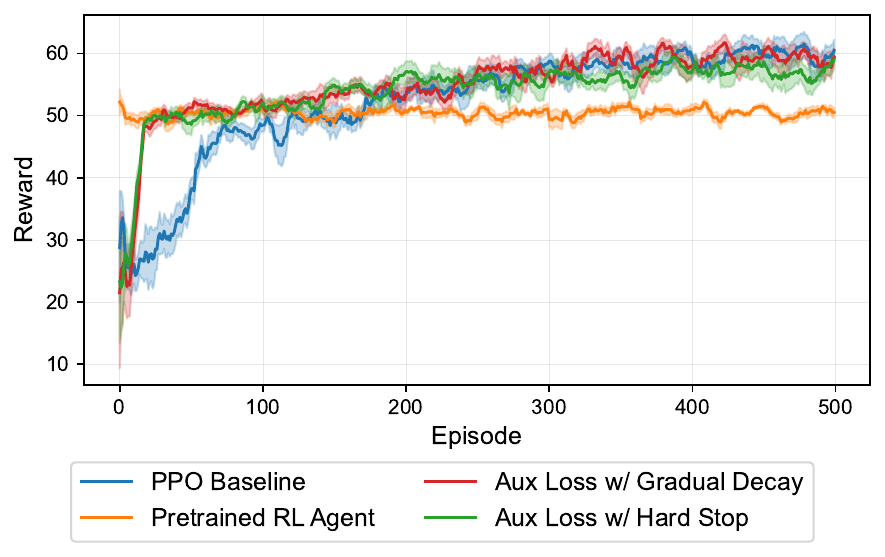}
    \caption{Comparing auxiliary loss against the PPO baseline across 10 independent runs of 500 episodes with a running average of 10. The shaded regions represent a ±1 SE.}
    \label{fig:auxlosseval}
\end{figure}

We show the auxiliary loss results in Fig. \ref{fig:auxlosseval}. The results indicate that both the decaying loss and hard stop implementations quickly converge to the teacher's performance by approximately episode 20, which is five time quicker than the baseline PPO agent, which does not reach the teacher's performance until around episode 110. By approximately episode 180, the baseline catches up to the teacher-guided agents.

\subsection{Feature Space Modification}
\begin{figure}[!t]
    \centering
    \includegraphics[width=\columnwidth]{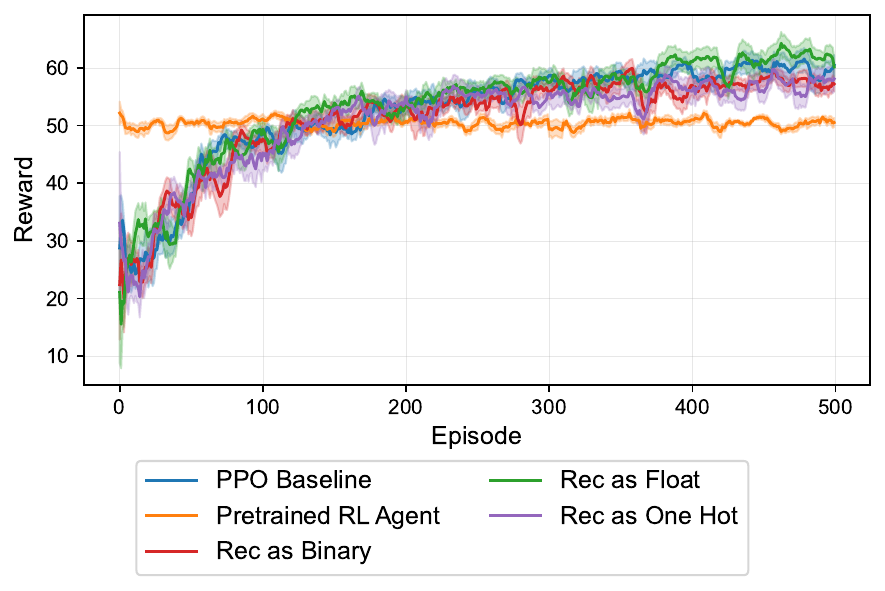}
    \caption{Comparing feature space modification against the PPO baseline across 10 independent runs of 500 episodes with a running average of 10. The shaded regions represent a ±1 SE.}
    \label{fig:featspaceeval}
\end{figure}

For feature space modification, we appended the teacher's recommendation in three forms: as a one-encoding, as a normalized float and as a binary value. Fig. \ref{fig:featspaceeval} presents the results of each technique. The results show no noticeable improvement for any of these techniques compared to the baseline. In fact, the teacher's guidance as a binary value yields a slightly lower policy by episode 300; however, this could be attributed to the stochasticity of the CybORG environment.

To verify whether the features corresponding to the teacher's recommendation had any impact on the agent's decisions, we used LIME to estimate the weight of the features for each technique. We present the results in Tables \ref{tab:featspacemodlime}a-c. Although Table \ref{tab:featspacemodlime}a shows that the teacher's one-hot encoded feature ranks second highest at episode 500, the teacher's recommendation is not included in the top four actions of the agent's policy - failing to demonstrate an ability to map the one-hot encoded recommendation to an executable action. Similarly, Table \ref{tab:featspacemodlime}b shows that although the teacher's recommendation had the highest probability at episode 500, the float encoded recommendation ranked 20th, and actually pushed the agent \textit{away} from selecting the teacher's recommendation.

For the binary encoding shown in Table \ref{tab:featspacemodlime}c, the teacher's recommendation is never among the agent's top 4 actions. Although individual features may have high weights, this fails to demonstrate any capability of the agent mapping these to a recommended action.

\begin{table}[!t]
\centering
\caption[LIME Results for Feature Space Techniques]{Using LIME to quantify the impact of features for the teacher's recommendation as (a) one-hot encoded; (b) a float; (c) a binary value. For (a), only the one-hot encoded feature weights are included. For (a) \& (b), \textit{Reco in Top 4} shows if the teacher's recommendation appears in the top 4 actions from the RL agent's policy, with its associated ranking if present. For (c), the ranking of each of the teacher's features is shown, with 1 having the highest impact on the agent's decision. Columns may be presented in an abbreviated form to keep the table concise.}
\small
\subfloat[Recommendation as One Hot Encoded]{
\resizebox{1\columnwidth}{!} {
\begin{tabular}{|c|c|c|c|c|}
\hline
\textbf{Episode}  &  \textbf{Weight}  &  \textbf{Ranking}  &  \textbf{Direction}  &  \textbf{Reco in Top 4} \\
\hline
1    &  1.22E-05   &  36  &  Towards  &  No \\
\hline
8    &  -6.90E-05  &  71  &  Away     &  No \\
\hline
16   &  2.78E-03   &  30  &  Towards  &  No \\
\hline
50   &  6.56E-03   &  18  &  Towards  &  No \\
\hline
100  &  2.66E-02   &  24  &  Towards  &  No \\
\hline
200  &  4.84E-02   &  38  &  Towards  &  No \\
\hline
300  &  3.13E-01  &  3   &  Towards  &  No \\
\hline
500  &  2.49E-01   &  2   &  Towards  &  No \\
\hline
\end{tabular}
}
}\\
\subfloat[Recommendation as Float]{
\resizebox{1\columnwidth}{!} {
\begin{tabular}{|c|c|c|c|c|}
\hline
\textbf{Episode}  &  \textbf{Weight}  &  \textbf{Ranking}  &  \textbf{Direction}  &  \textbf{Reco in Top 4} \\
\hline
1    &  -5.58E-06  &  40  &  Away     &  No     \\
\hline
8    &  -1.46E-04  &  33  &  Away     &  No     \\
\hline
16   &  -1.32E-03  &  25  &  Away     &  No     \\
\hline
50   &  -2.14E-04  &  44  &  Away     &  No     \\
\hline
100  &  -2.03E-03  &  44  &  Away     &  No     \\
\hline
200  &  1.00E-02   &  40  &  Towards  &  No     \\
\hline
300  &  -1.51E-02  &  35  &  Away     &  No     \\
\hline
500  &  -7.62E-02  &  20  &  Away     &  Yes/1  \\
\hline
\end{tabular}
}
}\\
\subfloat[Recommendation as Binary]{
\resizebox{1\columnwidth}{!} {
\begin{tabular}{|c|c|c|c|c|c|c|c|c|}
\hline
\textbf{Ep} & \textbf{F0} & \textbf{F1} & \textbf{F2} & \textbf{F3} & \textbf{F4} & \textbf{F5} & \textbf{F6} & \textbf{Reco in Top 4} \\
\hline
1   & 29 & \textbf{3}  &\textbf{ 1}  & \textbf{7}   & 52 & \textbf{5}  & \textbf{6}   & No \\
\hline
8   & \textbf{8}  & 47  & 13 & 25 & 22 & 36  & \textbf{9}   & No \\
\hline
16  & \textbf{3}   & \textbf{1}  & \textbf{8}   & 31  & \textbf{7}   & \textbf{9}   & 43 & No \\
\hline
50  & \textbf{4}  & 27  & 25 & 15  & 33  & \textbf{1}  & 16  & No \\
\hline
100 & \textbf{8}   & \textbf{6 } & \textbf{4}   & 20  & 22 & \textbf{3}   & 40  & No \\
\hline
200 & \textbf{1}  & 19  & \textbf{3}  & 36 & \textbf{10} & \textbf{5}   &\textbf{ 9}   & No \\
\hline
300 & 37 & \textbf{9}   &\textbf{ 7}   & 11 & 19 & \textbf{5}   & 17  & No \\
\hline
500 & 26 & 15  & 16  & \textbf{7}  & 19  & \textbf{2}   & 20  & No \\
\hline
\end{tabular}
}
}

\label{tab:featspacemodlime}
\end{table}

\subsection{Comparison}
We present the comparison of the teacher-guided techniques in Table \ref{tab:teachertechniquecomparison}. The two techniques that showed the most promising results were action masking, with its high initial performance, and incorporating the guidance as an auxiliary loss signal, with its quick convergence to the teacher's policy. Integrating the teacher's feedback through reward shaping and feature space modification showed no noticeable improvements in training.

\begin{table}[H]
\caption{Comparison of the teacher-guided techniques}
\begin{tabular}{|c|c|c|}
\hline
\textbf{Technique} & \textbf{Ranking} & \textbf{Notes} \\
\hline
Auxiliary Loss & 1 & Quick initial convergence \\
\hline
Action Masking & 2 & High initial performance \\
\hline
Reward Shaping & 3 & No noticeable improvement \\
\hline
Feature Space Modification & 4 & No noticeable improvement \\
\hline
\end{tabular}
\label{tab:teachertechniquecomparison}
\end{table}

\FloatBarrier
\section{Conclusion}
In this paper, we proposed improving the training efficiency of current ACO applications by integrating a teacher into the RL pipeline. We implemented and evaluated four distinct techniques within the CybORG environment, and showed that incorporating the teacher's guidance as an auxiliary loss signal and action masking yielded the most promising results, while reward shaping and feature space modification provided no measurable benefits. Our findings highlight the potential of teacher-guided RL for accelerating learning and improving early-stage performance for ACO.

This study used a pretrained RL agent to act as the teacher, requiring two rounds of training. Future work could explore integrating existing external knowledge sources, eliminating the need for additional training. Moreover, combining different teacher-guided techniques could further improve performance.

\bibliographystyle{unsrt}
\bibliography{skeleton}

\end{document}